\documentclass[letterpaper, 10 pt, conference]{ieeeconf}  
\usepackage{amsmath,amsfonts,amssymb}
\let\labelindent\relax
\usepackage{enumitem}
\usepackage{prettyref}
\usepackage{mathrsfs}
\usepackage{graphicx}
\usepackage{wrapfig}
\usepackage{subfloat}
\usepackage{MnSymbol}
\usepackage{multirow}
\usepackage{booktabs}
\usepackage[table]{xcolor}
\usepackage{cellspace}
\usepackage{mathtools}
\usepackage{sidecap}
\usepackage[noend]{algpseudocode}
\usepackage{comment}
\usepackage{duckuments}
\usepackage[ruled,vlined,linesnumbered]{algorithm2e}
\usepackage[colorlinks,allcolors=gray,hypertexnames=true]{hyperref}

\algnewcommand{\LineComment}[1]{\State \(\triangleright\) #1}
\algdef{SE}[DOWHILE]{Do}{doWhile}{\algorithmicdo}[1]{\algorithmicwhile\ #1}
\newcommand*{\colorboxed}{}
\def\colorboxed#1#{%
  \colorboxedAux{#1}%
}
\newcommand*{\colorboxedAux}[3]{%
  \begingroup
    \colorlet{cb@saved}{.}%
    \color#1{#2}%
    \boxed{%
      \color{cb@saved}%
      #3%
    }%
  \endgroup
}

\newrefformat{Fig}{Fig.~\ref{#1}}
\newrefformat{fig}{Fig.~\ref{#1}}
\newrefformat{par}{Section~\ref{#1}}
\newrefformat{appen}{Appendix~\ref{#1}}
\newrefformat{sec}{Section~\ref{#1}}
\newrefformat{sub}{Section~\ref{#1}}
\newrefformat{table}{Table~\ref{#1}}
\newrefformat{alg}{Algorithm~\ref{#1}}
\newrefformat{Alg}{Algorithm~\ref{#1}}
\newrefformat{Def}{Definition~\ref{#1}}
\newrefformat{Thm}{Theorem~\ref{#1}}
\newrefformat{Lem}{Lemma~\ref{#1}}
\newrefformat{step}{Step~\ref{#1}}
\newrefformat{ln}{Line~\ref{#1}}
\newrefformat{eq}{Eqn.~\ref{#1}}
\newrefformat{eqn}{Eqn.~\ref{#1}}
\newrefformat{pb}{Problem~\ref{#1}}
\newrefformat{it}{Item~\ref{#1}}
\newrefformat{te}{Term~\ref{#1}}
\def\Eqref Eq:#1:{\eqref{eq:#1}}
\newrefformat{Eq}{Equation~\Eqref#1:}

\newcommand{\E}[1]{\mathbf{#1}}
\newcommand{\TE}[1]{\textbf{#1}}

\newcommand{\argminP}[1]{\E{argmin}\;}

\newcommand{\argmaxP}[1]{\E{argmax}\;}

\definecolor{darkgreen}{HTML}{186a3b}

\usepackage[normalem]{ulem}
\usepackage{xcolor}
\definecolor{Blue}{rgb}{1,0,0}

\IEEEoverridecommandlockouts                              

\title{\Large\bf Tactile Probabilistic Contact Dynamics Estimation of Unknown Objects}
\author{Jinhoo Kim$^{*1}$, Yifan Zhu$^{*2}$, and Aaron Dollar$^{2}$
\thanks{$^*$: Denotes equal contributions.}%
\thanks{$^{1}$J. Kim is  with the Department of Mechanical Engineering, ETH Zurich, Zurich, Switzerland. Work done as a visiting scholar at Yale University. {\tt\small kimjin@ethz.edu}}%
\thanks{$^{2}$Y. Zhu and A. Dollar are with the Department of Mechanical Engineering and Materials Science, Yale University, New Haven, United States. {\tt\small  \{yifan.zhu, aaron.dollar\}@yale.edu}}%
}
\begin{document}
\thispagestyle{empty}
\pagestyle{empty}
\maketitle

\begin{abstract}
We study the problem of rapidly identifying contact dynamics of unknown objects in partially known environments. The key innovation of our method is a novel formulation of the contact dynamics estimation problem as the joint estimation of contact geometries and physical parameters. We leverage DeepSDF, a compact and expressive neural-network-based geometry representation over a distribution of geometries, and adopt a particle filter to estimate both the geometries in contact and the physical parameters. In addition, we couple the estimator with an active exploration strategy that plans information-gathering moves to further expedite online estimation. Through simulation and physical experiments, we show that our method estimates accurate contact dynamics with fewer than 30 exploration moves for unknown objects touching partially known environments.

\end{abstract}

\section{Introduction}
While robot manipulation technologies have advanced rapidly, deploying robots that can robustly perform manipulation with external contact in novel environments such as assembly and cooking remains a significant challenge. Model-free methods require environments that closely resemble or are identical to the training environment and degrade in the unfamiliar settings. Meanwhile, model-based approaches require dynamics models that are challenging to quickly acquire in unstructured environments where visual occlusions and sensor noises are prevalent. 

\begin{figure}[t!]
\centering
    \includegraphics[trim=0cm 7.7cm 7cm 2.2cm,clip,width=1\linewidth]{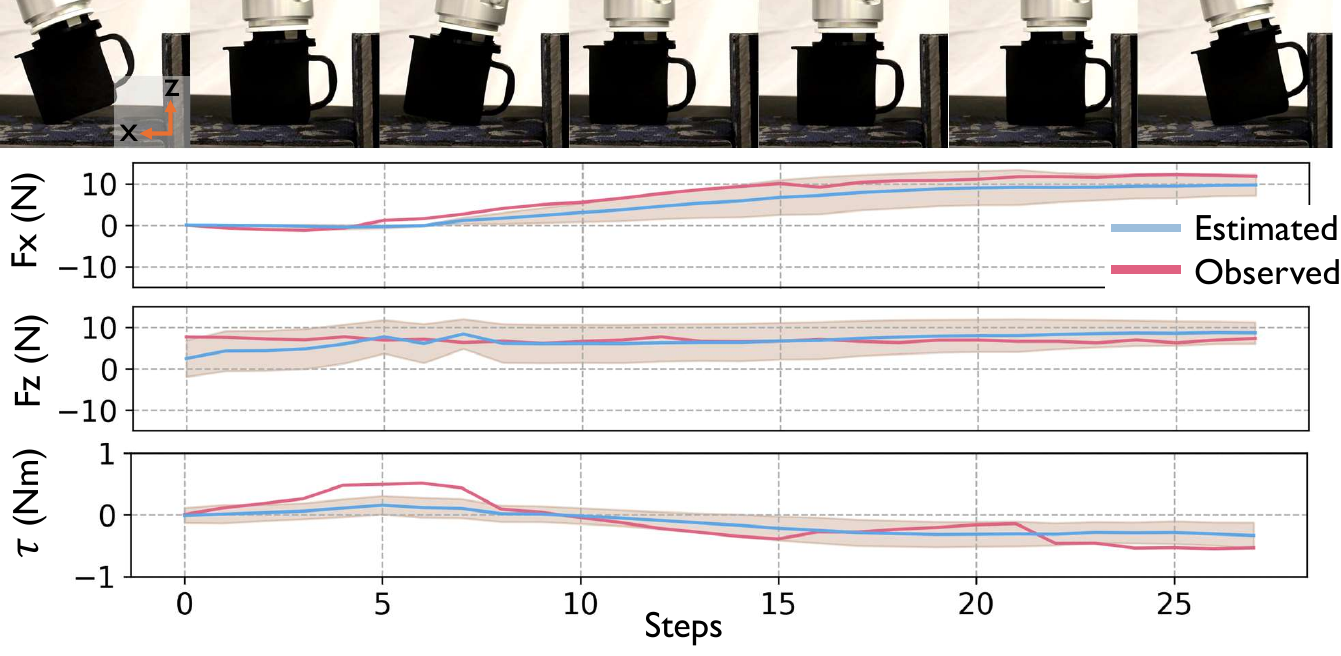}
    \caption{Our work considers estimating the contact dynamics of unknown planar rigid objects rigidly attached to a robot with an unknown rigid transform in a fully rigid partially known environment. Our method uses a particle filter with a compact representation of object geometry to quickly estimate the contact dynamics accurately based on tactile measurement only.}
    \label{fig:star_figure}
\end{figure}

In this work, we aim to rapidly identify an accurate contact dynamics model for a grasped unknown rigid object in a partially known rigid environment based exclusively on tactile measurements, and iteratively refine the model during interaction as shown in Fig.~\ref{fig:star_figure}. Previous works have focused on explicitly estimating the contact locations and types~\cite{doshi2022icra, taylor2023object} or obtaining a linear complementarity model~\cite{Jin2024}. However, explicitly determining contact positions and types is very challenging in real-world scenarios with complex object shapes, where there are abundant contact modes that also change frequently. In addition, a linear complementary model is a local approximation that inevitably introduces approximation errors especially when the local contact geometries are highly nonlinear. 

Instead, we contribute a novel formulation of the contact dynamics estimation problem as the joint estimation of contact geometries and physical parameters, and our proposed method quickly captures contact dynamics in the wild with few assumptions. Essential to our method is a compact and expressive geometric representation leveraging DeepSDF~\cite{park2019deepsdf}. This yields a representation of the object geometry using a learned continuous signed distance function (SDF) representation based on neural networks. The compact geometric representation allows us then to adopt a particle filter, which has shown robustness for tasks involving nonlinear and discontinuous contact dynamics~\cite{filter1,MPF}. Our representation ensures that the parameter space remains low dimensional, avoiding the curse of dimensionality of sampling-based methods, and enables the object parameters and geometry to be jointly estimated in a particle filter. In addition to the estimator, the quality of online samples also plays a key role in efficient estimation. We augment the particle filter with an active exploration strategy based on information theory that plans exploration actions with maximum expected information gain. 

We evaluate our method on unknown objects in an environment of unknown friction with a flat surface and a vertical wall of unknown height and position. In both simulation and physical experiments, our estimator quickly estimates the contact dynamics with high accuracy. In simulation, the estimator shows less than 4\,N of contact force prediction errors on new testing trajectories with ground truth force magnitudes going up to 25\,N, while in physical experiments, it predicts with less than 0.5\,N of error with 10\,N of ground truth force magnitudes. This is achieved after fewer than 30 exploration actions. In addition, in simulation, we show that the active exploration strategy reduced wrench prediction error by more than 10\% compared to random action selection.

\begin{figure*}[ht!]
\centering
    \includegraphics[trim=0cm 7cm 5.1cm 0cm,clip,width=0.8\linewidth]{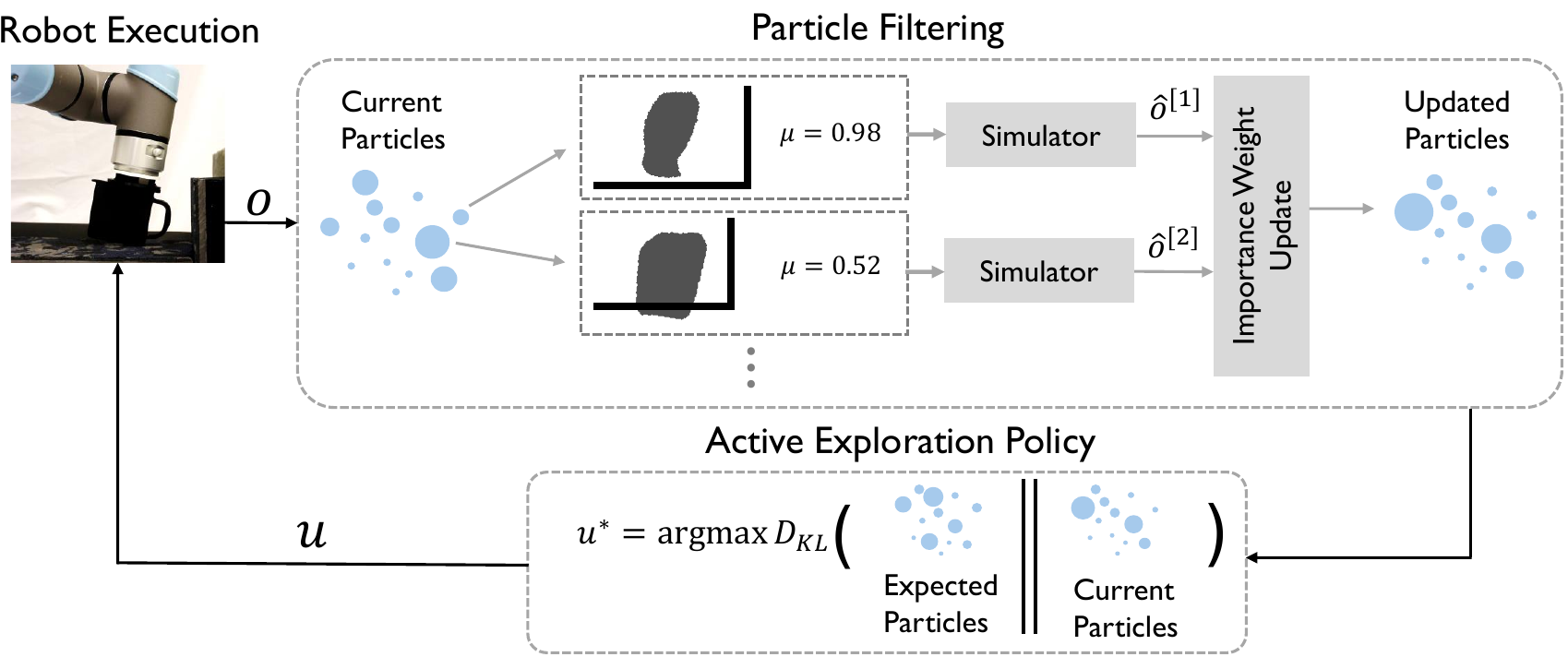}
      \vspace{-10px}
    \caption{Overview of the proposed estimation process. The robot manipulates an unknown object in a rigid environment with unknown surface height and wall position. Our method leverages DeepSDF~\cite{park2019deepsdf} (a compact learned SDF) as a representation of object geometry allowing joint estimation of geometry and physical parameters using a particle filter. We then employ an active exploration strategy to select actions which give the maximum expected information gain measured in terms of the Kullback–Leibler (KL) divergence. }
    \label{fig:method}
\end{figure*}

\section{Related Work}

\subsection{Contact Dynamics Estimation for Tactile Measurements}
Many existing works aim to explicitly estimate the contact type (e.g. point vs. line contact) and contact information (e.g location of point contact) from tactile measurements. These works adopt different estimation strategies, including direct equation fitting for narrow classes of contacts~\cite{doshi2022icra}, maximum a posterior (MAP) estimation via factor graphs~\cite{taylor2023object, kim2022active, kim2023simultaneous}, and particle filters~\cite{BayesianHybridCF, hertkorn2012identification, Pankert2023iros}. Some of these works assume unknown object geometry and a partially known environment, and estimate contact dynamics by explicitly considering the contact type, parameters, and the coefficient of friction~\cite{doshi2022icra, taylor2023object, kim2022active, kim2023simultaneous, BayesianHybridCF}. While these methods are effective when estimating limited contact types or modes, they fail to scale to cases with multiple contacts of different contact types or when the contact changes rapidly in realistic manipulation tasks. Other works~\cite{hertkorn2012identification, Pankert2023iros} assume that object and environment geometries are known. The most similar prior work to this was presented by Pankert and Hutter~\cite{Pankert2023iros}, where a particle filter is used to estimate the in-hand transform of an insertion object and the world transform of a target box of known geometries. The key difference is our method estimates these transforms along with the geometries and physical parameters with minimal prior knowledge, as is essential for deploying robots in novel unknown environments.  

Instead of directly estimating the contact parameters, another line of work aims to directly learn local dynamics, modeled as a linear complementarity system~\cite{pmlr-v168-jin22a,Jin2024}. In these works, the matrices of a linear complementarity system are fitted from observation data. Compared to these, our work estimates the geometries in contact, and avoids approximation errors from linearization.

\subsection{Particle Filtering for Contact Dynamics}
Particle filtering is a non-parametric filtering approach that can support multimodal probability distributions using particles. This property is highly valuable for rigid body contact estimation as the problem is often multi-modal. Many existing works have adopted particle filters for estimation problems involving contact or nonlinear dynamics~\cite{koval2013pose, ManifoldPF_SDF2017, filter1, filter2, MPF}. Our main contribution lies in the novel formulation of the contact estimation problem as nonlinear multimodal filtering of both geometry and physical parameters.

\section{Problem Definition}
We consider quasi-static manipulation of planar rigid objects rigidly attached to the robot with an unknown rigid transform, in a fully rigid environment. In this work, we assume the object geometry is unknown and the environment geometry is partially known, e.g. a flat surface with an unknown height and a wall located at an unknown position. Our goal is to estimate the probabilistic dynamics function $p(x_{l+1}, w_{l+1}|x_l, u_l)$ at time $l$, given past observations of the object poses $x_1, \cdots, x_l \in SE(2)$, contact wrenches $w_1, \cdots, w_l \in \mathbb{R}^3$, and position commands of the impedance controller $u_1, \cdots, u_l \in SE(2)$, . For convenience of notation, we also denote observation $o_l = [x_l, w_l]$. 


\begin{figure}[]
\centering
    \includegraphics[trim=0cm 12cm 9cm 0.1cm,clip,width=1\linewidth]{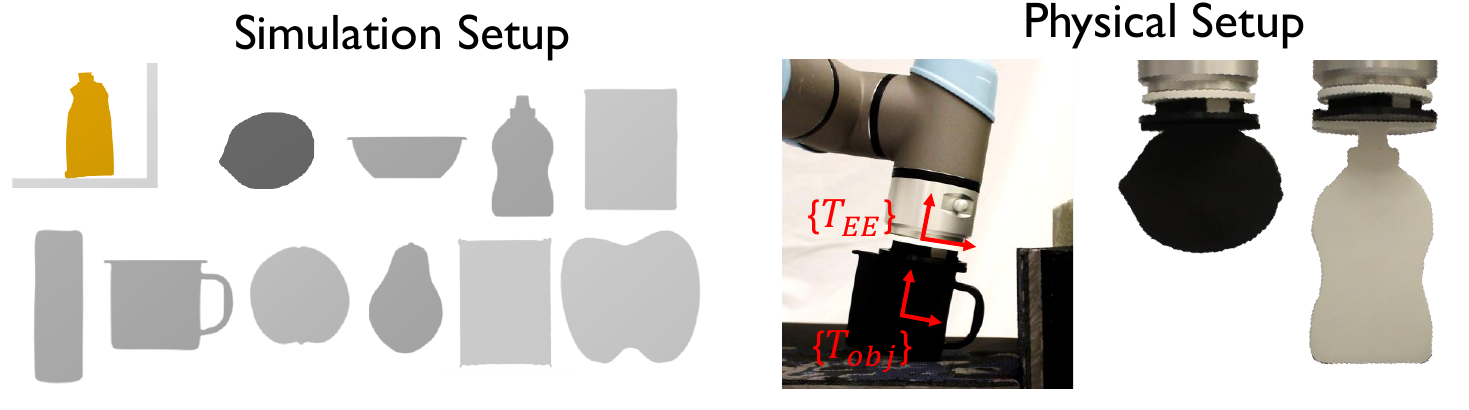}
      \vspace{-20px}
    \caption{Experiment Setup for both simulation (left) and physical world (right). In simulation, the environment used is a flat surface with an unknown height and a wall with an unknown position (Wall). The objects tested are 2D slices of YCB objects~\cite{ycb}. In physical experiments, the Mustard\_Bottle, Mug, and Lemon are tested in the Wall environment. Contact wrenches are measured by the internal end-effector force torque sensor (F/T) of the UR5e robot arm.  }
    \label{fig:experiment_setup}
\end{figure}
\section{Method}
We propose a method to efficiently capture the contact dynamics via the estimation of contact geometries and physical parameters with minimal assumptions. We take advantage of the well-established quasi-static rigid body contact dynamics and solve the problem of estimating the contact dynamics via the geometry and physical parameters $\theta$ of a quasi-static rigid body simulator. The dynamics function is then represented as $p(x_{l+1}, w_{l+1}|x_l, u_l) = \int p(x_{l+1}, w_{l+1}|x_l, u_l, \theta)bel(\theta)d\theta$, where $p(x_{l+1}, w_{l+1}|x_l, u_l, \theta)$ is the simulator and $bel(\theta)$ is the current belief of $\theta$, which is recursively updated over time using particle filter. In the case of a deterministic simulator, we simply denote the simulator as $(x_{l+1}, w_{l+1}) = f_\theta(x_l,u_l)$. Summarized in Fig.~\ref{fig:method}, our method first adopts a compact and expressive geometric representation in the form of a learned continuous Signed Distance Function (SDF) representation of the distribution of geometries. Then, treating contact dynamics estimation as a nonlinear multimodal filtering problem on the unknown parameters of a quasi-static rigid body simulator, we adopt a particle filter for efficient estimation. Finally, we couple our estimator with an active exploration strategy based on information theory to collect samples with high information. In this section, we will first discuss the geometric representation, then detail the quasi-static rigid body simulator, present our estimation algorithm based on particle filtering, and finally the active exploration algorithm.

\subsection{Geometry Representation}
A key ingredient of our method is the choice of the geometric representation. Representations such as meshes or point clouds, while extremely flexible, contain too many parameters for any practical objects, making the estimation intractable. Instead, we take advantage of an object geometry prior by utilizing DeepSDF~\cite{park2019deepsdf}, which is a signed distance function based on a neural network that can represent a broad class of shapes in a single latent vector. We denote it as $g(p,z):\mathbb{R}^2 \times \mathbb{R}^{d_z} \rightarrow \mathbb{R}$, which takes in the query 2D position of a point in the object frame $p \in \mathbb{R}^2$, and the latent geometry vector $z \in [-1,1]^{d_z}$, and outputs the signed distance $d \in \mathbb{R}$ at the query position from the object surface (positive for being outside of the object, and negative for being inside). We use the trained latent vector $z$ and the scaling parameter $s$ to represent the object geometry. 

For experiments, we assume that the unknown objects are drawn from the same distribution that the DeepSDF is trained on and we have partial knowledge of the terrain geometry, e.g. flat surface with unknown height. However, partial knowledge is not necessarily a limitation of our method since a similar compact geometric representation of the terrain to the object can be adopted and we intend to explore this in future work. As we will demonstrate in the Results section, we use $d_z = 2$ and achieved accurate estimation across all objects tested.


\subsection{Quasi-Static Contact Dynamics}
In quasi-static simulation, we assume that the robot moves slowly such that the Coriolis forces and accelerations can be ignored. We also assume the rigid bodies are always in force equilibrium, which is a mild assumption for many robot manipulation tasks. In addition, the robot uses a Cartesian-space impedance controller for commanding the end-effector poses with known impedance $K$. Note that due to the quasi-static assumption, there is no longer a concept of velocity, and the system is driven by position commands of the impedance controller of the robot. We follow the simulator proposed by Pang et al.~\cite{pang2021}, and briefly present the key aspects of the simulator below.

Denoting the state of the actuated and un-actuated rigid objects in the environment with subscripts $a$ and $un$, at each time step, the quasistatic rigid body simulator solves a linear complementarity problem (LCP) at time $l$:
\begin{align*}
    &\text{Find $v_{l+1}$, s.t.:} \\
    &J_{c,u}^Tf_{c,un} + h\tau_{un}= 0 \\
    &J_{c, a}^Tf_{c,a} + h\tau_a + hK(x^* - x_l - hv_{l+1})= 0 \\
    &0 \leq \phi_l + hf_c^Tv_{l+1} \perp f_c \geq 0\\
    & \text{Terrain constraints}.\\
\end{align*}
Where $v_{l+1}$ is the velocity at the next step and $h$ is the time step such that $x_{l+1} = x_l + hv_{l+1}$. Note that quasi-static systems do not actually have velocity, this term here is just used for simulating the objects forward in time. $J_c$ is the contact Jacobian, $f_c$ is the frictional contact force that depends on the coefficient of friction $\mu$, and $\tau$ are external forces including gravity and constraint forces such as those that constrain a rigid object that is a fixed terrain to be static. $\phi$ is the signed distance function. The expression $hK(x^* - x_l - h*v_{l+1})$ is the impulse applied by the impedance controller of stiffness $K$ over $h$. Note that we use the minus sign $-$ for both positional and angular differences for rotation matrices. In our problem, we consider robots with a single rigidly attached actuated object interacting with static terrain. This LCP can be expressed as the KKT condition of a quadratic program~\cite{pang2021}, and we solve it efficiently with OSQP~\cite{osqp}.

\subsection{Contact Dynamics Estimation via Particle Filtering}

\begin{algorithm}
\caption{Particle Filtering Step}
\KwIn{Current Particle Set $\Theta_l$ and Their Associated Importance Weights $\Omega_l$, $O_{l}$,  $U_{l}$, $R$, $r$ }
\For{i $\leftarrow$ 1 \KwTo M}{
    $\omega^{[i]}_{l+1} \leftarrow 1$ \\
    \For{j $\leftarrow$ 1 \KwTo N -1}{
     $o,o', u \leftarrow O_l[j], O_l[j+1], U_l[j]$ \\
    $\hat{o} = f_{\theta_l^{[i]}}(o, u) $\Comment{Predict Observation} \\
    $\omega^{[i]}_{l+1} \leftarrow \omega^{[i]}_{l+1}  \omega^{[i]}_l \mathcal{N}(\hat{o}; o', R)$ \Comment{Update Weights}}
}
$\Omega_{l+1} \leftarrow \text{Weight Normalization}$ \\
\If{$n_{eff}(\Theta_t) \leq \frac{M}{2}$}
{$\Theta_{l+1} \sim Bel(\Theta_l)$, $\Omega_l = \frac{1}{M}$ \Comment{Resample}\\
$\Theta_{l+1} \leftarrow \Theta_{l+1} + \beta$, $\beta \sim \mathcal{N}(0, rVar(\Theta_{l}))$ \Comment{Roughening}
}
\KwRet{$\Theta_{l+1}, \Omega_{l+1}$ }
\label{alg:pf}
\end{algorithm}

We treat dynamics estimation as a nonlinear multimodal filtering problem on the geometry and physical parameters $\theta$. Therefore, $\theta$ is the concatenation of the latent SDF shape vector, the size scale, the pose of the object with respect to the end-effector, the environment shape parameter (such as the position of an unknown wall), and the coefficient of friction $\mu$. Given the nonlinear and discontinuous nature of contact dynamics, we adopt particle filter, a non-parametric filtering approach that has shown good performance for problems involving nonlinear dynamics~\cite{filter1, filter2, hang2021manipulation}. In particle filters, we denote the particle set with $M$ particles at time $l$ as: $\Theta_l \coloneq \theta_l^{[1]}, \cdots, \theta_l^{[M]}$ and the associated weights at time $l$ as: $\Omega_l \coloneq \omega_l^{[1]}, \cdots, \omega_l^{[M]}$. The belief of the state at time $l$, or in our case $Bel(\Theta_l)\approx \sum^M_i{\omega^i_l \delta (\theta_l - \theta_l^i)}$, is represented as a set of particles $\theta_l^i$ and the associated weights $\omega_l^i$. The essence of the particle algorithm is the same as the Bayes filter but with the beliefs represented as particles. This allows the particles to represent multi-modal distributions, which is very common for contact dynamics estimation. 

Our particle filtering algorithm is summarized in Alg.~1. Given the current belief of particles, it first predicts the observation at time $l$ based on the previous observation and the previous command input.  Then, the importance weight of each particle is calculated by the probability density function of a Gaussian centered at the observation with a fixed diagonal covariance matrix $R \in \mathbb{R}^{d_\theta}$~\cite{particle_filter_tutorial}, after which the weights are normalized. One challenge in using tactile feedback for contact dynamics estimation is that a single measurement is not very informative. To overcome this challenge, instead of only using the most recent observation and action, we use a history of $N$ recent observations and actions (denoted as $O_l$ and $U_l$) to update $Bel(\Theta_l)$, which was first introduced in~\cite{Memory}. Here, we note that particles are not updated through a process model as the particles represent the fixed geometry and physical parameters. Instead, we use the roughening method \cite{roughening} by adding artificial noise after resampling to prevent particle depletion, which happens when a small number of particles dominate the distribution. Artificial noise is sampled from a zero-mean Gaussian with variance scaled by a roughness $r$ relative to the particle variance. To also mitigate particle depletion due to frequent resampling~\cite{particle_filter_tutorial}, we only resample when the effective sample size $n_{eff}=\sum^M_i\frac{1}{\omega^2_i}$ is $\leq M/2$.


\subsection{Active Exploration}
\begin{algorithm}
\caption{Expected Information Gain}
\KwIn{Action $u$, $\Theta_l$, $\Omega_l$}
$G \leftarrow 0$ \Comment{Initialize EIG} \\
\For{i $\leftarrow$ 1 \KwTo M}{
    $\hat{o}^{[i]}_{l+1} = f_{\theta_l^{[i]}}(x_l, u_l)$ \Comment{Simulate the Observation} \\
    $\Bar{\Theta}_l, \Bar{\Omega}_{l}\leftarrow$ Particle Filter Step with Alg.~1 \\
    $G \leftarrow G + w_l^{[i]}D_{KL}(\Bar{\Theta}_l,\Bar{\Omega}_l \Vert \Theta_l, \Omega_l)$ \Comment{Calculate Information Gain} \\
}
\KwRet{$G$}
\label{alg:EIG}
\end{algorithm}

Instead of executing random exploration moves to collect information, we adopt an active exploration strategy based on information theory. We follow the active learning approach for particle filter proposed by Hauser~\cite{hauser2010randomized}. We choose an action that would maximize the expected infomration gain (EIG), where the information gain is defined as the Kullback–Leibler (KL) divergence between the belief of $\theta$ before and after an update from the observation $o$. Specifically, we choose an action $u^*$ from a set of action candidates $U$ according to:
\begin{align*}
    &u^* = \underset{u \in U}{\text{argmax}}\,\text{EIG}(u) = \\
    &\qquad \int_{o} D_{KL}(Bel(\theta|o)||Bel(\theta))P(o|\theta, u)do\\
    &D_{KL}(Bel(\theta|o)||Bel(\theta)) = \int_{\theta}Bel(\theta|o)log\frac{Bel(\theta|o)}{Bel(\theta)} d\theta\\
\end{align*}

To calculate EIG for a single action, we take advantage of the particles representation of $Bel(\theta)$. Shown in Alg.~2, for each particle, we simulate the observation with the action $u$, perform a particle filter step (without resampling) based on the observation, and calculate the KL divergence for the updated belief. Then EIG is the weighted sum of these information gains. Note that the KL divergence can be easily calculated by directly using the weights before and after the importance weight updates. This is an $O(NM^2)$ operation that requires many calls to the simulator where $N$ is the number of actions~\cite{hauser2010randomized}. Therefore, we randomly downsample M/5 particles and weights for EIG calculation. 


\section{Experiments and Results}

\begin{figure}[ht!]
\centering
    \includegraphics[trim=0.5cm 3.5cm 9.6cm 8.5cm,clip,width=1\linewidth]{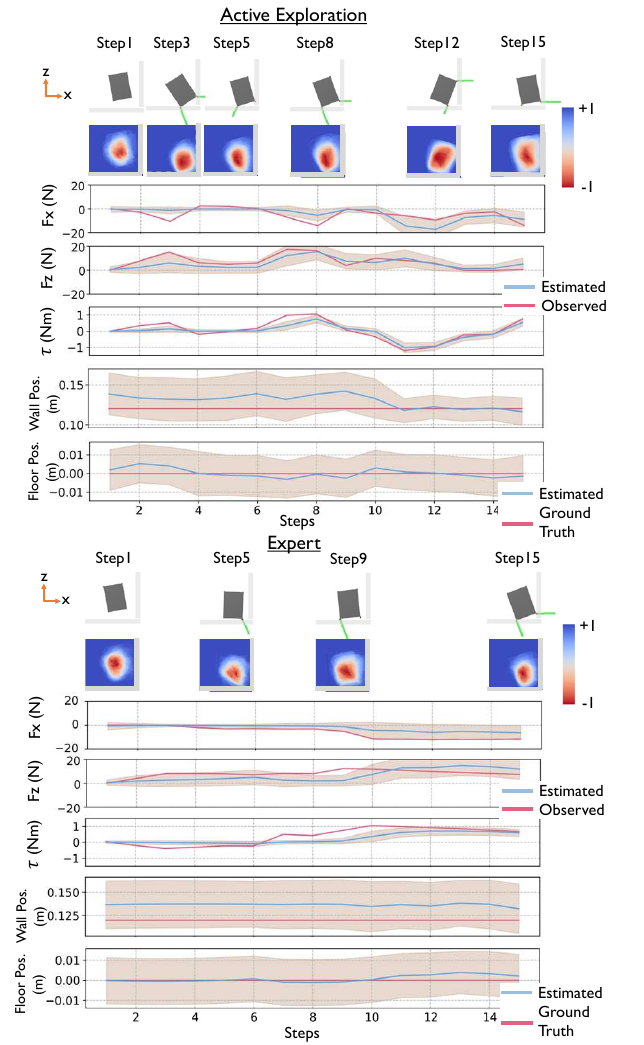}
      \vspace{-20px}
    \caption{Estimation progress for the Master\_Chef\_Can object for Active exploration and Expert exploration policies. The estimations of wrench, wall position, and floor position during the exploration trajectory are plotted with standard deviation. The trajectories and the current distribution of geometries are also visualized, with contact forces shown as green lines. The geometry distribution is the weighted indicator function ($+1$ outside of the object and $-1$ inside the object) for the top 100 particles in $\Theta$ with the largest weights.}
    \label{fig:simulated_estimation}
\end{figure}

\begin{figure}[h!]
\centering
    \includegraphics[trim=0cm 14cm 6cm 3.3cm,clip,width=1\linewidth]{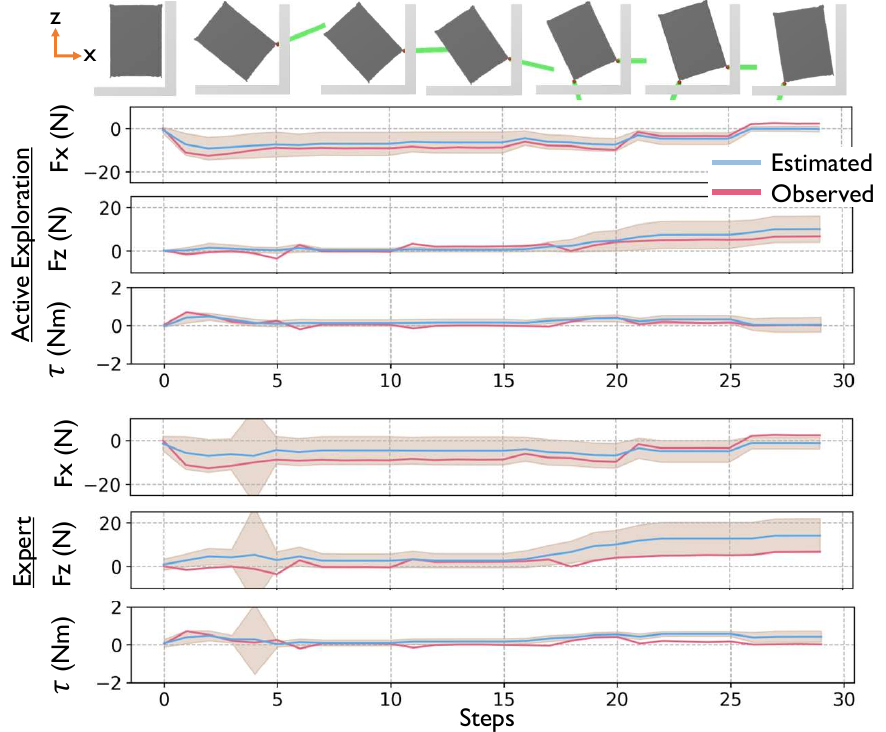}
      \vspace{-20px}
    \caption{Evaluation of the dynamics model estimated by Active exploration, and Expert exploration policies for the Master\_Chef\_Can object. The estimations of wrenches are plotted with standard deviation. The evaluation trajectory is also visualized at the top.}
    \label{fig:simulated_eval}
\end{figure}

\begin{table*}[]
    \renewcommand{\arraystretch}{1}
    \centering
    \footnotesize
    \begin{tabular}{@{}ccccccc@{}}
        \toprule
         {\bf Environment}& {\bf Exploration Strategy} & {\bf$F_x (N)$}  & {\bf $F_z(N)$} & {\bf $\tau (Nm)$}& {\bf $x (mm)$}& {\bf $z (mm)$}\\  
        \midrule
        \multirow{ 3}{*}{Sim} &Random & $3.71\pm3.92$ & $3.23\pm4.11$ & $0.260\pm0.619$ & $7.33\pm10.1$ &$6.00\pm4.52$ \\ 
        &Active  & $3.66\pm2.24$ & $2.81\pm2.14$ & $\mathbf{0.212\pm0.182}$ &  $7.72\pm 10.0$ &$5.64\pm4.05$ \\ 
        &Expert & $\mathbf{3.14\pm1.70}$ &$\mathbf{2.68\pm2.74}$ & $0.213\pm0.140$ & $\mathbf{5.87\pm9.49}$ &$\mathbf{5.40\pm4.05}$ \\ 
        \midrule
        Real &Expert & $0.490\pm0.688$ & $0.212\pm1.378$ & $0.111\pm0.235$ & $5.88\pm4.15$ &$1.90\pm13.4$ \\ 
        \bottomrule
    \end{tabular}
    \caption{Testing results for simulation and real experiments, averaged over 11 objects on 3 testing trajectories and 3 objects on 1 testing trajectory, respectively.}
    \vspace{-10px}
    \label{tab:results}
\end{table*}

We evaluate our estimation pipeline in both simulation and physical experiments. In this section, we first present the results of simulated experiments and then discuss physical experiments. We train the DeepSDF function on the 2D cross-sections of selected 21 objects from the YCB dataset~\cite{ycb} to represent the object geometry, where the latent geometry vector dimension $d_z = 2$. For both simulation and physical experiments, we set the impedance for the controller as $K = [100\,\text{N/m},100\,\text{N/m},50\,\text{Nm}]$. Note that the controller is very stiff in rotation and compliant in $x$- and $z$-axis. In addition, each action changes the target pose to approximately translate 1\,cm or rotate $5^\circ$.


\begin{figure*}[h!]
\centering
     \includegraphics[trim=1cm 24.6cm 1cm 0cm,clip,width=1\linewidth]{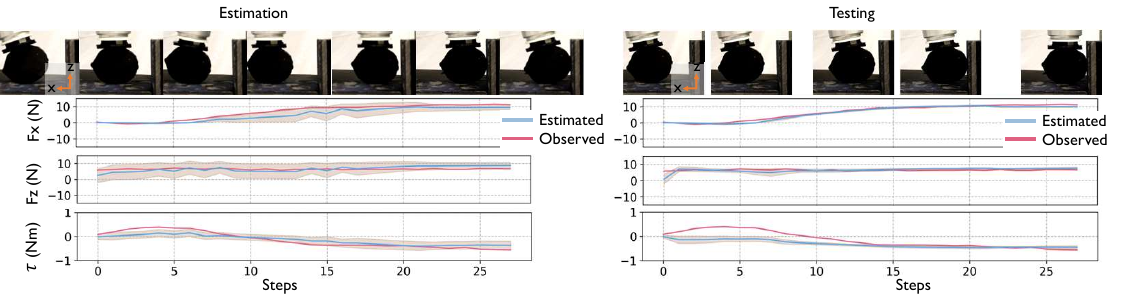}
     \vspace{-25px}
    \caption{Example trial of the Lemon object in physical experiments. Left: The estimation progress by our estimator, with an expert exploration trajectory. Right: Testing of the estimated dynamics model on a different trajectory, where both the estimated wrench and the standard deviation are shown.}
    \label{fig:physical_results}
\end{figure*}
\subsection{Simulated Experiments}
Shown in Fig.~\ref{fig:experiment_setup}, our environment in simulation is a flat ground with unknown height $g_h$ and a vertical wall located at an unknown position $p_w$. The parameter $\theta$ is a 9-dimensional vector, including the DeepSDF latent vector $z$, the object pose relative to the end-effector frame $T_{o,ee}$, size scale $s$, and the surface coefficient of friction $\mu$. We assume that we know the possible range of the unknown parameters, where the unknown parameters are selected from a uniform distribution of $g_h\sim\mathcal{U}[-0.02m, 0.02m]$,  $p_w\sim\mathcal{U}[0.09\,\text{m}, 0.18\,\text{m}]$, $z\sim\mathcal{U}[[-1.0,-1.0], [1.0,1.0]]$, $T_{o,ee}\sim\mathcal{U}[[-0.02\,\text{m}, -0.02\,\text{m}, -0.2\,\text{rad}], [0.02\,\text{m}, 0.02\,\text{m}, 0.2\,\text{rad}]]$, $s\sim\mathcal{U}[0.8, 1.2]$, and $\mu\sim\mathcal{U}[0.1, 0.9]$.  

As shown in Fig.~\ref{fig:experiment_setup}, the simulation experiments use 11 planar objects which are cross sections of objects from the YCB dataset. For the particle filter, we use $M = 5,000$ particles, the observation noise parameters of $[R_{F_{x}}, R_{F_{y}}, R_{\tau}, R_{x}, R_{z}, R_{\phi}] = $ $[30\,\text{N},30\,\text{N},0.3\,\text{Nm},0.0001\,\text{m},0.0001\,\text{m},0.002\,\text{rad}]$, roughness of $r=0.3$, and memory length $N=5$. For each object in each environment, we compare three different exploration strategies, examples of which are shown in Fig.~\ref{fig:simulated_estimation}. The first is a baseline exploration strategy that adds random movements to a basic exploration trajectory (Random). This basic exploration trajectory simply commands the robot to make contact with the floor and move to the right. A zero mean Gaussian noise with variance $[0.03 \text{m}, 0.03 \text{m}, 0.25 \text{rad}]$ is added to the position command. The second is our active exploration strategy which adds exploration actions to the same basic exploration trajectory. The action set is 27 position commands that is a uniform grid with range $[\pm0.03\text{m}, \pm0.03\text{m}, \pm0.025\text{rad}]$. Note that due to limitations in computation, our active exploration strategy only plans one step ahead. Therefore, this is a local strategy and it requires the basic exploration trajectory as guidance to avoid getting stuck. We hope to alleviate the computation requirement of our active exploration strategy in the future. The last is an ``expert" exploration trajectory which is tuned by the authors based on the ground truth wall position. We use the same expert strategy across all the objects. We run each exploration strategy for 15 time steps. 

Once the estimation is completed, we fix the particles and test the estimated dynamics model on three different testing trajectories in the same partially known environment for 30 time steps. One example is shown in Fig.~\ref{fig:simulated_eval}. Presented in Table~\ref{tab:results}, we report the quantitative results, averaged over 11 objects on 3 testing trajectories. The metric we used are the mean absolute error (MAE) between the ground truth and the weighted mean predictions of the top 100 particles with the largest weights. We report MAE for wrench predictions and next pose predictions except for rotation. This is because we use a very stiff controller in rotation and the error is minuscule. Overall, active exploration outperforms Random, but is worse than the Expert trajectory. We think that this is mainly due to the short horizon and the small action set we are using due to computation limit. Meanwhile, we observe that for the particular example shown in Fig.~\ref{fig:simulated_eval}, active exploration outperforms Expert. This shows that a fixed exploration strategy, as is done for Expert, is not necessarily good for all objects, demonstrating the value of active exploration. As shown in Fig.~\ref{fig:simulated_estimation}, active learning approach uses both the bottom and top right part of the object to make a contact with the wall to simultaneously estimate unknown object and environment geometry. As a result, from time step 11, the wall position prediction quickly converges to the ground truth position. In terms of computation, it takes less than 1\,s to do the estimation step and about 5\,s to perform an active exploration step on a computer with an Intel i9-13900KF CPU, 64 GB of RAM, and an NVIDIA GeForce RTX 4090 GPU.

\subsection{Physical Experiments}
As shown in Fig.~\ref{fig:experiment_setup}, we use a UR5e robot arm with 3D printed objects mounted on the end effector to perform the task with three objects in the Wall environment. We implemented an impedance controller on the UR5e robot. As the UR5e does not offer torque control, we approximate impedance control by using the position controller and the end-effector F/T sensor. However, this requires the environment to be deformable. We used gym tiles that deform about 1-3\,mm with a 10\,N contact force in our experiments. Despite this violation of the rigid body assumption, we show that our estimator still performs accurate estimation. 

We use the same hyperparameters for the particle filter, except for the observation noise parameter $R = $ $[20\,\text{N},20\,\text{N},0.5\,\text{Nm},0.1\,\text{m},0.1\,\text{m},0.1\,\text{rad}]$. Here, the initial distribution over the unknown parameters are the following: $g_h\sim\mathcal{U}[-0.02\,\text{m}, 0.02\,\text{m}]$,  $p_w\sim\mathcal{U}[-0.90\,\text{m}, -0.80\,\text{m}]$, $z\sim\mathcal{U}[[-1.0,-1.0], [1.0,1.0]]$, $T_{o,ee}\sim\mathcal{U}[[-0.02\,\text{m}, -0.02\,\text{m}, -0.2\,\text{rad}], [0.02\,\text{m}, 0.02\,\text{m}, 0.2\,\text{rad}]]$, $s\sim\mathcal{U}[0.8, 1.2]$, and $\mu\sim\mathcal{U}[0.5, 1.6]$. To showcase the best performance of our estimator, instead of using the active learning strategy, we adopt an Expert exploration trajectory for these experiments.  We report the quantitative results in Table~\ref{tab:results} and show the estimation and testing trajectories for Lemon in Fig.~\ref{fig:physical_results}. The estimation of Mug is also shown in Fig.~\ref{fig:star_figure}. Despite the only information given in this case is that the shape of the environment is flat with a wall, our estimator is able to quickly estimate the contact dynamics, achieving less than 0.5\,N of force prediction error where the ground truth magnitude goes up to 10\,N. We believe our method shows great promise towards contact dynamics estimation in the open world.

\section{Discussion and Conclusion}
In this work, we present a method to quickly estimate accurate contact dynamics for unknown objects in a partially known environment. Through both simulated and physical experiments, we demonstrate the accuracy of our estimator. We also show the effectiveness of our active exploration approach in the simulated experiments. One requirement of our method is the presence of a good geometry prior. We believe that with the abundance of 3D geometry data and the development of 3D large vision foundation models, such a requirement is not an obstacle.

There are a number of future directions we would like to pursue. First, we would like to extend this to 3D and lift the restrictions for a partially known environment. We hope to also investigate techniques to improve the sample efficiency of particle filters. Next, we would like to improve the computation efficiency of our active exploration strategy. Additionally, we want to explore training a reinforcement learning agent to learn an exploration strategy that best suits our estimation pipeline. Finally, we would like to adopt our estimator for downstream manipulation tasks. 


\bibliographystyle{IEEEtran}
\newpage
\bibliography{references}
\end{document}